\documentclass[14pt]{article}

\usepackage{arxiv}
\usepackage{authblk}

\usepackage{times}
\usepackage{epsfig}
\usepackage{graphicx}
\usepackage{amsmath}
\usepackage{amssymb}
\usepackage{booktabs}

\usepackage{subcaption}
\usepackage{url}  
\usepackage{amsfonts}
\usepackage{bm}
\usepackage{caption}
\usepackage{algorithm}
\usepackage{algpseudocode}
\usepackage{multirow}
\usepackage[table,x11names,svgnames]{xcolor}

\def\wacvPaperID{1427} 

\usepackage[pagebackref=true,breaklinks=true,colorlinks,bookmarks=false]{hyperref}

\begin{document}

\title{Conviformers: Convolutionally guided Vision Transformer}
\date{}


\author[1,2,3,*]{\bf \large Mohit Vaishnav}
\author[1,2]{\bf \large Thomas Fel}
\author[2]{\bf \large Iva\'n Felipe Rodr\'iguez }
\author[1,2]{\bf \large Thomas Serre}
\affil[1]{Artificial and Natural Intelligence Toulouse Institute, Universit\'e de Toulouse, France}
\affil[2]{Carney Institute for Brain Science, Dpt. of Cognitive Linguistic \& Psychological Sciences,
Brown University, Providence, RI 02912 }
\affil[3]{Centre de Recherche Cerveau et Cognition, CNRS, Universit\'e de Toulouse, France} 
\affil[*]{Corresponding author: \href{mohit_vaishnav@univ-toulouse.fr}{mohit\_vaishnav@univ-toulouse.fr}}

\maketitle
\thispagestyle{empty}

\begin{abstract}

Vision transformers are nowadays the de-facto choice for image classification tasks. There are two broad categories of classification tasks, fine-grained and coarse-grained. In fine-grained classification, the necessity is to discover subtle differences due to the high level of similarity between sub-classes. Such distinctions are often lost as we downscale the image to save the memory and computational cost associated with vision transformers (ViT). In this work, we present an in-depth analysis and describe the critical components for developing a system for the fine-grained categorization of plants from herbarium sheets. Our extensive experimental analysis indicated the need for a better augmentation technique and the ability of modern-day neural networks to handle higher dimensional images. We also introduce a convolutional transformer architecture called \textit{Conviformer} which, unlike the popular Vision Transformer (ConViT), can handle higher resolution images without exploding memory and computational cost. We also introduce a novel, improved pre-processing technique called \textit{PreSizer} to resize images better while preserving their original aspect ratios, which proved essential for classifying natural plants. With our simple yet effective approach, we achieved SoTA on Herbarium 202x and iNaturalist 2019 dataset. 

\end{abstract}

\section{Introduction}
\label{sec:intro}

Artificial Intelligence is about to transform botany by facilitating the identification of both extant and extinct species -- including a novel, unknown species. One of the first attempts to automate leaf classification includes the work by \cite{wu2007leaf} using a probabilistic neural network \cite{specht1990probabilistic} which calculated decision boundary following the Bayes strategy. An early attempt to use a convolutional architecture for plant classification includes the work of \cite{yalcin2016plant,islam2019patanet}. A pre-processing step was required in \cite{yalcin2016plant} to tackle the challenges posed by the dataset, such as blurring and illumination changes. Later, a multi-layered perceptron (MLP) based approach was tested by  \cite{pacifico2018plant}. Most algorithms during the early stage followed two steps, feature extraction and classification. \cite{swain2012approach} used information such as sepal length, sepal width, petal length, and petal width as a feature extractor and used it in a multi-layered feed-forward network for IRIS plant identification.   Unlike the approaches mentioned above, modern deep learning methods \cite{Wu2019,wang2021feature,chou2022novel,he2022transfg,zhang2022free} used the images to extract the features and do the classification because of the availability of increasingly large image datasets such as iNaturalist~\cite{van2018inaturalist} and Herbarium~\cite{herbarium2021}.

\begin{figure}[ht]
\centering
\includegraphics[width=.6\linewidth]{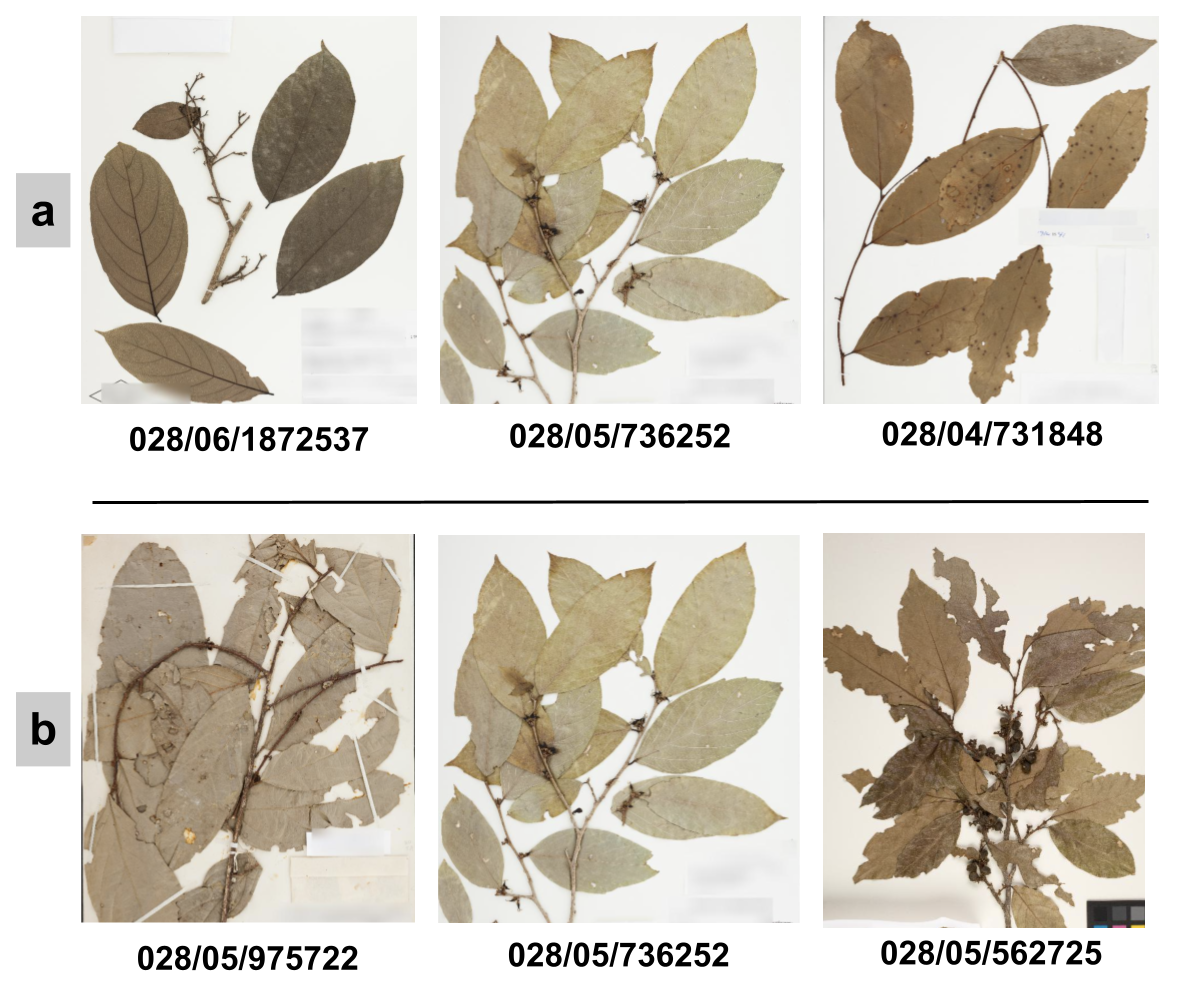} 
\caption{\textbf{Herbarium 2021 Dataset:} (a) Three representative examples from different classes showing the visual similarity. (b) Three representative examples from the same classes display intra-class variability. Respective identifiers are mentioned below each example.}
\vspace{5mm}
\label{fig:fgexample}
\end{figure}

     

Vision Transformers are one widely accepted class of architectures that have taken over the field of computer vision  \cite{carion2020end,Dosovitskiy2020-iq,d2021convit,strudel2021segmenter}. These architectures rely on multiple blocks of self-attention modules and feed-forward networks to compute the interaction between all the pairs of input tokens passed to the network. Their ability to model content-based interaction is a computationally intensive task and is quadratically associated with the length of the input sequence. To avoid memory explosion and the cost of training such transformer architectures, input images are usually downscaled before passing. However, this downscaling operation negatively impacts more challenging tasks, such as those presented in fine-grained classification. Unlike coarse-grained classification, sub-classes in a fine-grained classification task possess threefold challenges. First, there exists a very high level of similarity between species of different sub-class (Figure~\ref{fig:fgexample}(a)); second, there is high variation between samples of the same sub-class (Figure~\ref{fig:fgexample}(b)); and lastly, often, these datasets are limited in samples because of the need for a professional to annotate the data. Previous work has already shown that leaf classification, for instance, requires high-resolution images (see \cite{wilf2016computer}for instance).

Here, we propose a novel plug-and-play module for vision transformer architectures that augments its ability to handle higher resolution input images with insignificant computational cost. The need to process high-resolution images is especially problematic for transformer networks because of their large memory footprint, which grows as a function of the image resolution. Our proposed architecture effectively blends the benefits of convolutional with transformer-based architectures. Initial studies \cite{xiao2021early} have found the positive impact of merging convolutional layers with transformer networks. We also demonstrate the need for higher resolution images for better classification. A secondary contribution of this work includes a novel resizing technique (\textit{PreSizer}) to preserve the aspect ratio of the herbarium images -- which proved to be necessary for the fine-grained recognition of plant species. Our comprehensive experimental analysis shows how critical ingredients like regularization, dropout and hierarchical labels alter the model's performance and proposes a classification recipe to be followed.



\paragraph{Our contributions are as follows:}
\begin{itemize}
\itemsep0em
    \item A novel \textit{Conviformers} architecture that combines a convolutional module as a front-end with a vision transformer module as a back-end. 
    \item A generic approach that allows the transformer models to attend to input images of any resolution with minimal memory requirement.
    \item Empirical evidence exhibiting the need for higher image resolution to achieve better classification accuracy.
    \item A recipe after analyzing multiple components and their contribution to learning suitable discriminative features for classification.
    \item A novel pre-processing technique \textit{PreSizer} to resize images preserving the aspect ratio and improving recognition accuracy for fine-grained recognition.
    \item A top entry uni-model performance on the Herbarium-202x and iNaturalist 2019. 
\end{itemize}

\section{Related Work}
Four broad categories of research work incorporate convolutions with transformer architectures. First, inserting few attention modules in between residual blocks \cite{Wang_2018_CVPR,chen20182,yue2018compact,Shen_2021_WACV,vaishnav2021understanding}; second, inserting attention modules at the end \cite{huang2019ccnet,carion2020end,sun2019videobert}; third, augmenting convolution layers with attention modules in parallel \cite{bello2019attention}; and lastly substituting convolution layers by attention modules \cite{ramachandran2019stand,Zhao_2020_CVPR,Wang_2018_CVPR,hu2019local}.

Still, there persisted a wide gap between the learnability of vision transformers and CNNs. \cite{wu2021cvt,guo2022cmt,yuan2021incorporating,graham2021levit,dai2021coatnet,Peng_2021_ICCV} analysed the potential weaknesses in directly applying transformers from NLP and proposed a combination of transformers along with convolutional networks. \cite{wu2021cvt} proposed Convolutional vision Transformers (CvT) and presented convolutional-based patch projection of image tokens along with hierarchical design. Another alternative, LocalViT \cite{li2021localvit} proposed depthwise convolution to capture local features; Levit \cite{graham2021levit} on the other hand, enhanced the inference speed of ViT by first passing the images from a series of convolutional blocks and passing its output to the ViT. A network proposed by \cite{zhou2021elsa} incorporated locality without convolutions with the help of enhanced local self-attention using Hadamard attention and ghost head. Hadamard attention enabled maintaining higher-order mapping considering neighboring attention, while ghost heads increased the channel capacity by combining attention maps.  

A vanilla transformer model learns attention maps independently at each layer. To improve the quality of the attention maps, evolving attention transformers were proposed in \cite{wang2021evolving}. A chain of residual convolutional modules is used to guide attention mechanisms to interact between the maps in different layers. However, all these architectures require the input image to be smaller in resolution because of the increasing computational complexity associated with increasing input resolution. 

Recently researchers have focused on a fifth direction where they focus on augmenting attention layers soft convolutions inductive bias. \cite{d2021convit} replaced the first few self-attention layers of ViT \cite{Dosovitskiy2020-iq} with gated positional self-attention to mimic the locality of convolutional layers. Their architecture uses a gating parameter to control how much to emphasize spatial positions vs. image content. This architecture starts with a convolutional layer as a default using the gates, which feed into self-attention modules. We base our work on this model because of its sample efficiency in learning and the ability to switch from attention to convolutions. Unlike other convolutional augmented transformer architectures proposed in the literature, we propose a plug-and-play convolutional module in this work. This module can be attached as a front-end to any vision transformers.

Transformer are also gaining popularity in fine-grained classification tasks \cite{wang2021feature,hu2021rams,he2022transfg,zhang2022free}. This classification field can be divided into localization methods and feature encoding methods. In the localization \cite{ge2019weakly,liu2020filtration,yang2021re} method, the focus is to localize discriminative areas of the image that are later used for classification. Whereas in the feature encoding \cite{yu2018hierarchical,zheng2019learning,gao2020channel} method, the focus is on learning relevant higher-order features. Vision transformers backbones have been used in localizing discriminative regions in methods like FFVT \cite{wang2021feature}, TransFG \cite{he2022transfg}, RAMS-Trans \cite{hu2021rams} and AF-Trans \cite{zhang2022free}. A plug-in module,  PiM \cite{chou2022novel}, can be used either with the convolutional network or with ViT, where the emphasis is given to the strength of the attention maps instead modulated features map for classification. All of these networks work on a fixed input resolution input image. On the contrary, we propose a general purpose design that removes the constraints of working with fixed image resolution with transformer architecture.

\section{Dataset}
\label{sec:data}
\paragraph{iNaturalist 2019:} This dataset is composed of 268,243 images containing 1,010 classes at the species level. As we move from the coarse level, there are 3 classes for Kingdom, 4 classes for Phylum, 9 classes for Class, 34 classes for Order, 57 classes for the family, 72 classes for genus and 1,010 classes for Species. The iNaturalist \cite{van2018inaturalist} dataset is a long-tailed large-scale real-world
datasets. For a fair comparison, we used official splits of training and validation images in this paper.

\paragraph{Herbarium 2021:} Herbarium sheets are a vital source of botanical research. This dataset contains information about taxon, families and orders for 2.5M images. Herbarium sheets provided in the dataset belong to 451 families and 81 orders. At the finest level, there are 64500 species from the Americas and Oceania. This long-tailed dataset contains 3 images per species on the lower bound and $>$100 as the upper bound. It is a part of a project spearheaded by the New York Botanical Garden. This dataset \cite{herbarium2021} is a split training and test set with an 80\%/20\% ratio. We further divided the training dataset into a training and a test split with a 90\%/10\% ratio. We did not use the novel data augmentation technique mentioned in Section~\ref{sec:dp} while training the model on Herbarium 2021. 

\paragraph{Herbarium 2022:}

All experiments described here use the Herbarium 2022 dataset\footnote{\url{https://www.kaggle.com/c/herbarium-2022-fgvc9/data}}. It contains 15,501 species of vascular plants from North America, collected across 60 different botanical institutions worldwide. The distribution of class labels across the 1.05M images is long-tailed. There are as few as 7 samples per taxa and as many as 100 for some taxa. Training, validation and test distribution are the same as above. 

In addition to class labels, additional meta-data is available for the family, genus and species. The family and genus level hierarchy has a unique name. In contrast, at the species level, they are unique at their genus level, and there could be similar species names for the different genera. There are 272 families and 2564 genera in the Herbarium-2022 dataset. This dataset is skewed at all three levels and has a long-tailed distribution. 

\section{Data prepossessing}
\label{sec:dp}
Images in the Herbarium 2022 dataset have different resolutions, with their longest side always 1000 pixels. Working with such higher dimensional images is also computationally expensive. One possible widely accepted solution is to either resize the images or use padding to make them symmetric. However, we introduce a technique, \textit{PreSizer}, that maintains the aspect ratio of the images in the dataset and makes their resolution symmetrical. Our first observation was that most images have very distinctive borders and contain irrelevant information for the prediction. So we remove the boundary pixels of a width of 20px from all four sides. Next, we find the shorter edge of the image, and along this side, we pad it to make it equal to the longer side by reflecting the image. Following this approach makes our images of the resolution 1000$\times$1000 as the longer side is always 1000.

\begin{figure}[ht]
\centering
\includegraphics[width=.122\linewidth]{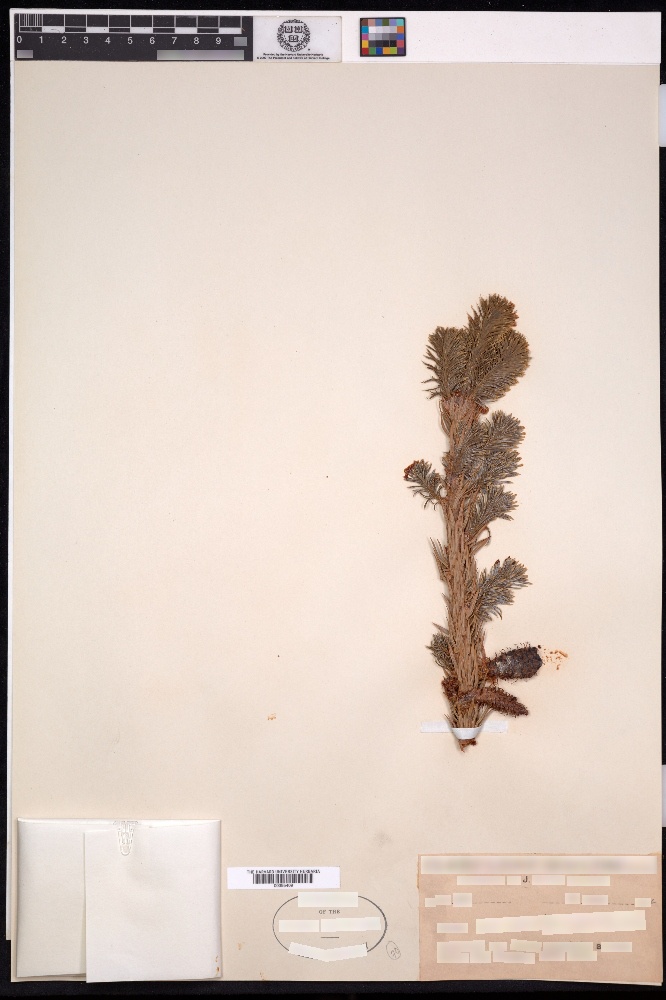} 
\includegraphics[width=.182\linewidth]{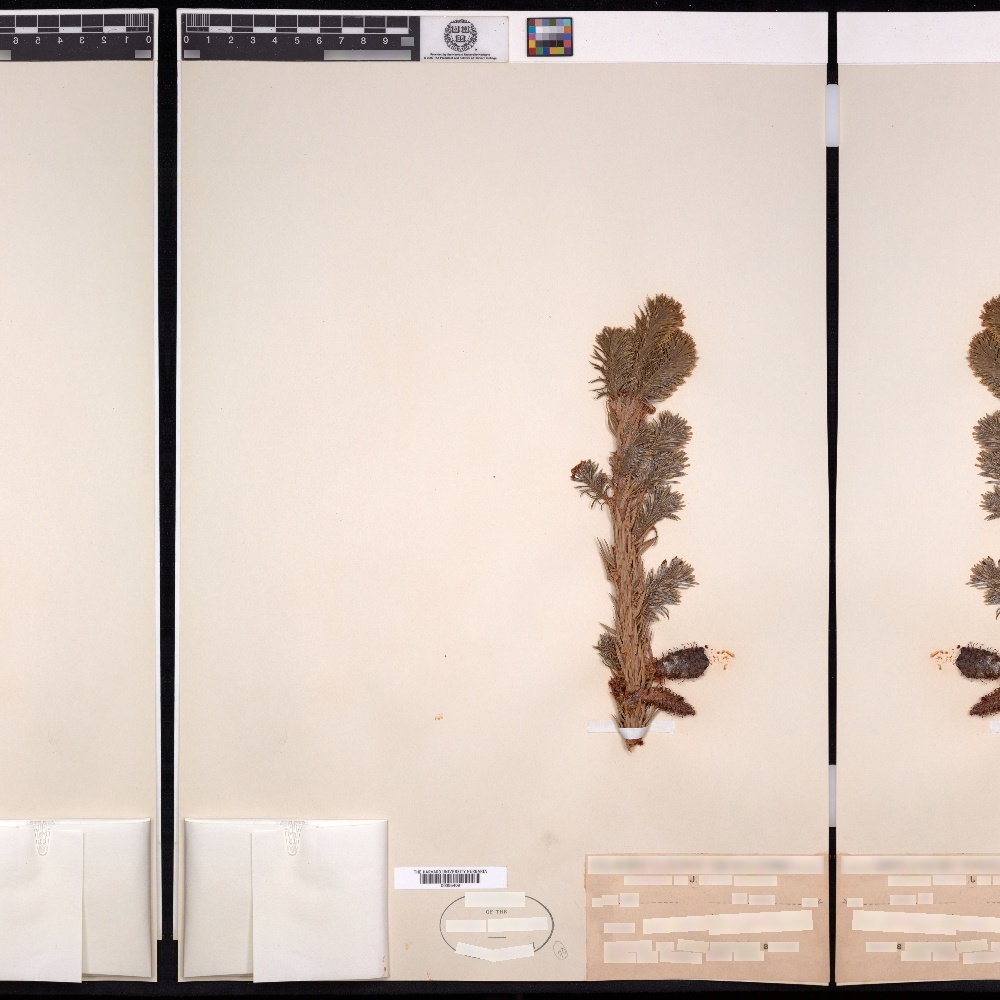} 
\includegraphics[width=.182\linewidth]{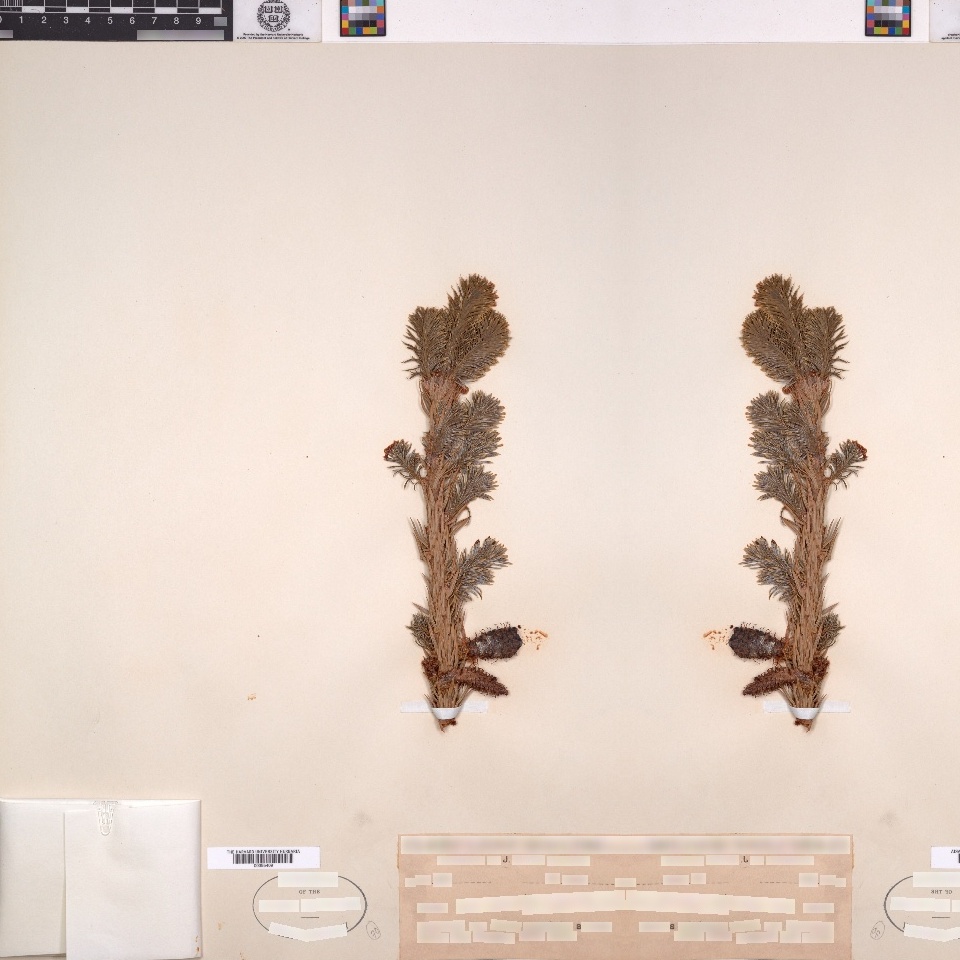} \\
\begin{subfigure}{.122\linewidth}
    \includegraphics[width=1\linewidth]{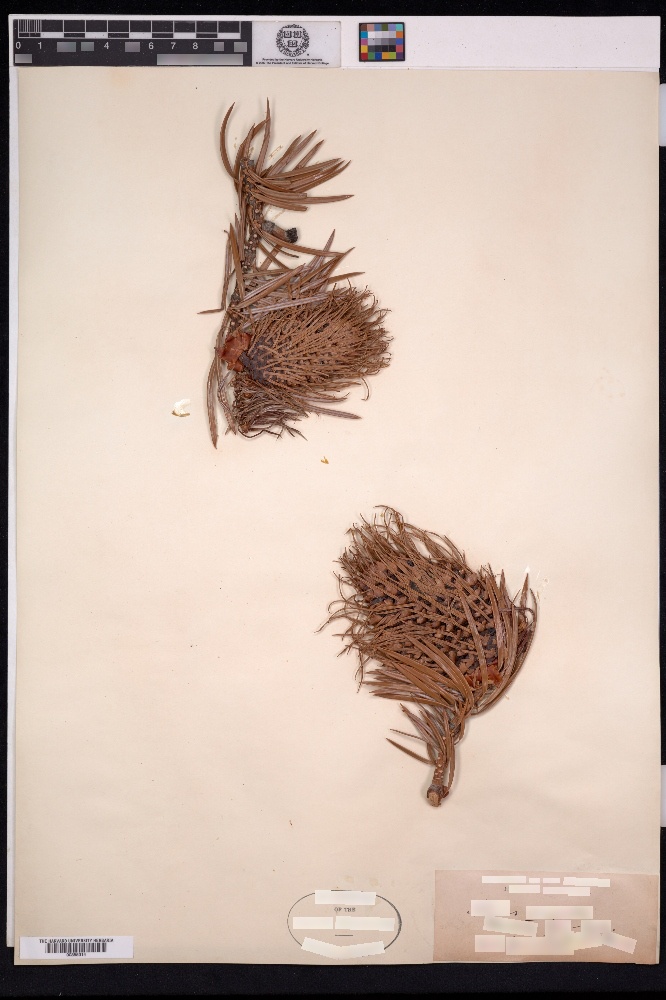}
    \caption{Reference}
\end{subfigure}
\begin{subfigure}{.182\linewidth}
    \includegraphics[width=1\linewidth]{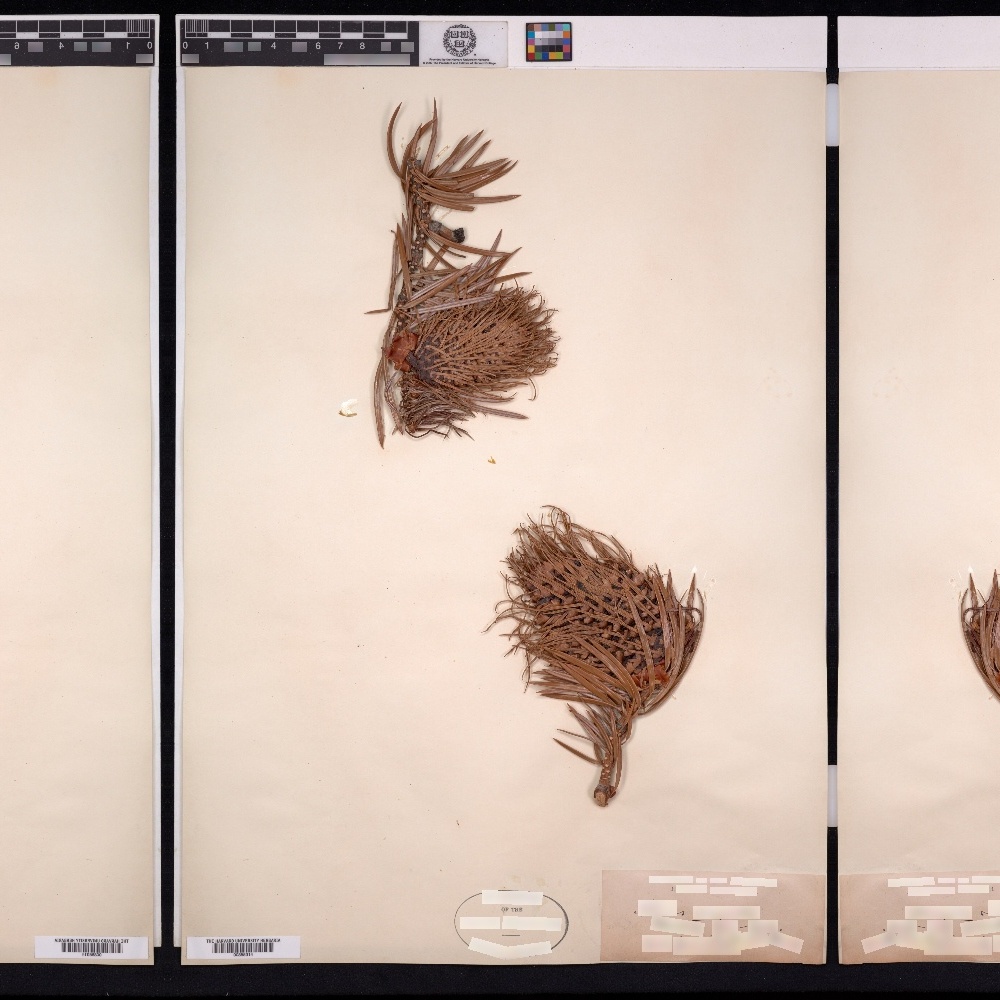}
    \caption{Equal padding }
\end{subfigure}
\begin{subfigure}{.182\linewidth}
    \includegraphics[width=1\linewidth]{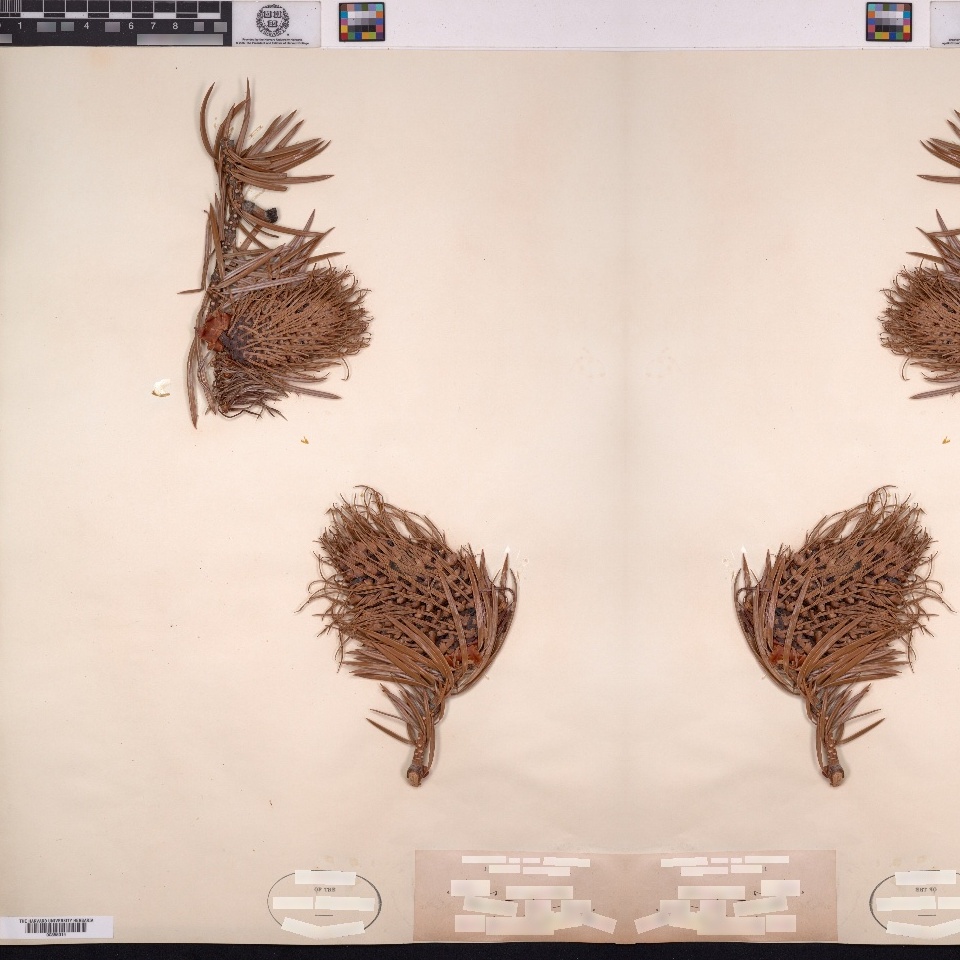}
    \caption{PreSizer}
\end{subfigure}
\caption{Two representative examples showing resizing-preserving. \textbf{(a)} Initial image from the Herbarium dataset. \textbf{(b)} Resizing the image when padding equally on both sides. When we follow equal padding around the sides, we lose relevant content. \textbf{(c)} Resizing the image following the  \textit{PreSizer} technique in which we remove the boundary pixels of width 20 pixels from all the sides and append the mirror reflection of the image to make it square.}
\vspace{5mm}
\label{fig:smartresize}
\end{figure}

We use these resized and augmented images with contextually relevant information for the training at a resolution of 512$\times$512 and central cropping to 448$\times$448. Figure~\ref{fig:smartresize} shows an illustrative example using \textit{PreSizer} strategy. As we can see, the images are smooth around the edges removing artifacts where we carry out reflection and do not introduce any discontinuity. We also used the Scikit-Learn \cite{scikit-learn} pre-processing library to encode the different labels in this given dataset.

\section{Conviformers}

\begin{figure}[ht]
\centering
\includegraphics[width=.6\linewidth]{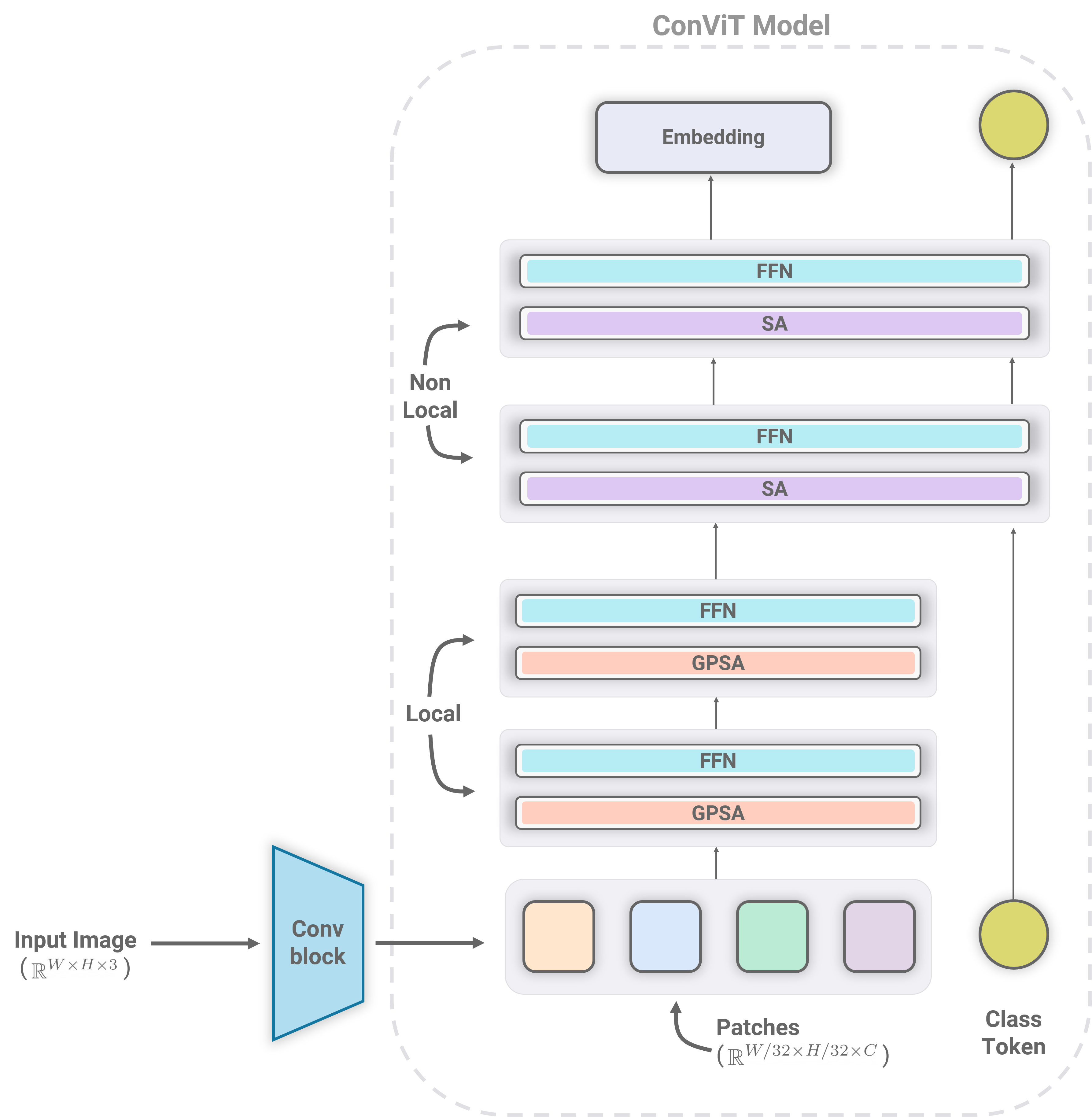} 
\caption{\textbf{\textit{Conviformer}}: Convolutionally guided vision transformer model. We selected the base architecture as ConViT and added a convolutional frontend to reduce memory demands with increasing image resolution.}
\label{fig:modelc3}
\vspace{5mm}
\end{figure}

We proposed an architecture \textit{Conviformers} (Figure~\ref{fig:modelc3}). It combines ConViT architecture that can perform at par or even better with standalone transformer \cite{Dosovitskiy2020-iq,touvron2021training} architectures in a lower data regime. ConViT tries to get the best from convolutions as well as transformers. It is initialized with a soft convolutional inductive bias using a learnable parameter which can also learn to switch to self-attention if needed. A gated positional self-attention (GPSA) learns a parameter $lambda$ controlling the balance between convolutionally initialized content-based attention and self-attention. This gating parameter pays less attention to the convolutional position throughout the training than self-attention. However, the early layers of the model still use convolutional inductive bias to help in training.

\textbf{Architecture}: ConViT require the input image divided in $14\times 14$ non-overlapping patches of 16$\times$16$\times$3. These patches are embedded in $D_{emb}$ = $64N_h$ dimensional vector using convolutional stem with $N_h$ as number of heads in the model. These patches are passed through 10 GPSA layers with convolutional initialization and two self-attention layers, followed by a two-layer Feed-Forwards network with GELU activation. To predict the input class ConViT uses an extra class token similar to BERT \cite{devlin2018bert}, appended after the last GPSA layer. We use the base variant of the model, which has 16 heads ($N_h$) and a 768-dimensional embedding vector ($D_{emb}$).

\begin{figure}[ht]
\centering
\includegraphics[width=.3\linewidth,angle=0]{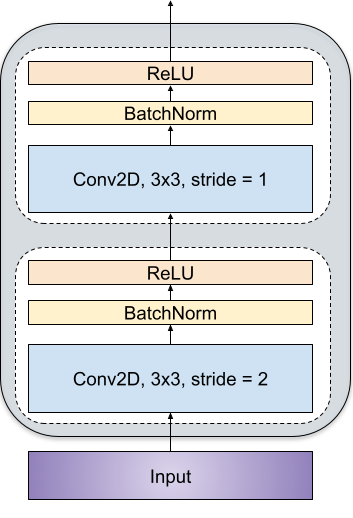} 
\vspace{5mm}
\caption{\textbf{\textit{Conviformer}}: Convolutionally block proposed in the model.}
\label{fig:model_conv}
\end{figure}

We add a convolutional block (Figure~\ref{fig:model_conv}) on top of ConViT architecture, enabling it to attend to images of larger dimensions. This convolutional block take the input image ($x$) of dimension $h\times$$w\times$$3$ (usually $h$ is equal to $w$)and downsampling if by a factor of ($d_s$) where 
\[ d_s = \lfloor h/224 \rfloor\ = \lfloor w/224 \rfloor\] and convert the image to $h^{'}$ $\times $ $w^{'}$ $\times$64
\[ h^{'} = \lfloor h/d_s \rfloor\ and ~~ w^{'} =  \lfloor w/d_s \rfloor\]

\[ total ~~patches~(t_p) = \frac{h^{'}}{16} \times \frac{w^{'}}{16} \]

This feature map ($z \in \mathbb{R}^{B\times t_p \times 64}$; $B$ is the batch size) is then sent to the subsequent attention blocks of ConViT, where the model's regular mechanisms continue to classify the images. In Table~\ref{tab:num_patch} we compare the number of patches needed to be processed while using ConViT with \textit{Conviformer}.

\begin{table}[]
\centering
\vspace{5mm}
\begin{tabular}{lcccc}
\hline
\multicolumn{1}{l}{\multirow{2}{*}{\textbf{Architecture}}} & \multicolumn{4}{c}{\textbf{Image Resolution}} \\ \cline{2-5} 
\multicolumn{1}{c}{}  & \multicolumn{1}{c}{224} & \multicolumn{1}{c}{448} & \multicolumn{1}{c}{512} & \multicolumn{1}{c}{600} \\ \hline
ConViT  &   196  & 784  &  1024   &  1369    \\
\textit{Conviformer}  & -  &  196  &   256     &  324  \\ \hline      
\end{tabular}
\vspace{5mm}
\caption{Number of patches (Sequence length) for different input image resolutions. Sequence length is directly proportional to the memory demand by the self-attention module.}
\vspace{5mm}
\label{tab:num_patch}
\end{table}

Similar to other Vision transformer architectures where different models are defined as \textit{tiny, small} and \textit{base}, we also use the same nomenclature. Unless otherwise specified we have refer \textit{Conviformer} to architecture built on \textit{base} model of \textit{ConViT}. 

\section{Approach}

\subsection{Resolution}
Our first experiment demonstrates the importance of input image resolution for a fine-grained classification task. For this experiment, we selected four resolutions (224$\times$224, 448$\times$448, 512$\times$512 and 600$\times$600). We trained ConViT for the smallest resolution image and \textit{Conviformer} for the rest. All the models are trained from scratch for 300 epochs with default hyperparameters. 

\subsection{Herbarium-2021} 
We first trained our model on the Herbarium-2021 dataset. We divided the dataset into training and validation, with 2M samples for training and 200k samples for validation purposes. We let the model train for 300 epochs with cross-entropy loss and default hyperparameters as mentioned on their GitHub\footnote{\url{https://github.com/facebookresearch/convit}}. We did not use \textit{PreSizer} as described in section \ref{sec:dp} in this training process. 

\subsection{Herbarium-2022} 
We used the pre-trained weights of Herbarium 2021 to initialize the model for current training instead of random initialization. The impact of pre-training granularity is discussed in prior works \cite{yan2020clusterfit,cui2019measuring}. Our model is trained in three stages. In the first stage, we trained the model with 90\% of the training dataset for 300 epochs. In the second stage, we included all the training datasets available for another 10 epochs. Finally, we modified the loss function with triplet loss and trained the model for 10 more epochs. Our model used images after they were passed through the \textit{PreSizer} and resized to 512$\times$512 and center cropped to 448$\times$448. We trained the model for 320 epochs, this time using their default hyper-parameters.  

\paragraph{Loss:} We use three other losses to train the model initialized with the pre-training on the previous year's Herbarium-2021 challenge. The first model is trained using \textit{cross-entropy} (CE) loss (equation~\ref{eq:ce}) at the taxa level for the first 300 epochs. Once this training is complete, we took \textit{triplet} loss \cite{triplet}  as a regularization technique to train the model for another 10 epochs further. Triplet loss (eq.~\ref{triplet_loss}) helps us to bring together the samples of the same taxa in their embedding space and separate different taxa apart from each other. 

\begin{equation}
    \label{eq:ce}
    \mathcal{L}_{CE} = \sum_{n=1}^N -w_{y_n}\log \frac{exp(x_{n,y_n})}{\sum_{c=1}^C exp(x_{n,c})}
\end{equation}

where $x$ is the input, $y$ is the target, $w$ is the weight, $C$ is the number of classes, and $N$ is the minibatch dimension.

\begin{equation} \label{triplet_loss}
\begin{split}
    \mathcal{L}_{\text{Trip}}( (\bm{x}, \bm{x}^{+}, \bm{x}^{-}) ; \bm{g} ) = 
    \text{max} ( \alpha & +
    || \bm{g}(\bm{x}) - \bm{g}(\bm{x}^{+}) ||_2^2
    - || \bm{g}(\bm{x}) - \bm{g}(\bm{x}^{-}) ||_2^2, 0)
\end{split}
\end{equation}

where $\bm{x}$ is an anchor sample, $\bm{x}^+$ is another instance from the same class as the anchor, while the negative $\bm{x}^-$ is from a different class than the anchor. The final loss function becomes $\mathcal{L}_{CE} + \mathcal{L}_{Trip}$.


\paragraph{Hierarchical labels}
This dataset has an enormous amount of relevant information which can further be used for classification, such as hierarchical labels for each sample image. These labels provide meta information related to family, species, and category labels. In Herbarium-2022, there are 272 types of families and 2564 types of genera in the dataset. Metadata also contains information about the phylogenetic distances between two genera. We used this information and explored the model's performance in different scenarios. 

In the first set of experiments, we used the two hierarchical labels, i.e., family and genus and trained the model with cross-entropy (CE) loss to better regularize the architecture. To do the same, we pass the extracted features of the network to the taxonomical classification layer $f(x)$ and also to the two multilayered perceptrons (MLP) output layers serving as the classification layer for the family $h(x)$ and genus labels $g(x)$ (equation~\ref{eq:hier}). 

\begin{equation} 
\label{eq:hier}
    \begin{split}
    \centering
        out_{feat} &= model(x) \\
        label_{tax} &= f(out_{feat}) \\
        label_{gen} &= g(out_{feat}) \\
        label_{fam} &= h(out_{feat}) 
    \end{split}
\end{equation}

where $out_{feat} \in \mathbb{R}^{768}$, $label_{tax}$ $\in \mathbb{R}^{15501}$, $label_{gen} \in \mathbb{R}^{2564}$ and $label_{fam} \in \mathbb{R}^{272}$. Loss for this model is calculated as shown below (equation \ref{ce_loss}). 

\begin{equation}
    \label{ce_loss}
    \mathcal{L} = \mathcal{L}_{CE_\text{tax}} +  
    \lambda_1 \mathcal{L}_{CE_\text{gen}} +   
    \lambda_2 \mathcal{L}_{CE_\text{fam}}
\end{equation}

In the next set of experiments, we added more regularization to the network. For this trial, we added triplet loss in addition to the cross entropy loss of the hierarchical labels (equation~\ref{eq:hiertrip}). We used two sampling techniques to draw samples at family or genus levels. For example, in genus-level sampling, positive samples are from the same genus labels as anchor labels, and negative is different from those. Similarly, we draw the same family label sample for a positive triplet at a family level, but it could have a different genus label than the anchor label. We added more output branches to the network in order to obtain the embedding features at the taxa ( $emb_{tax} \in \mathbb{R}^{512}$), genus ($emb_{gen} \in \mathbb{R}^{128}$) and family ($emb_{fam} \in \mathbb{R}^{128}$) level. These variables are considered in calculating loss during the training as mentioned in equation \ref{trip_loss}. 

\begin{equation} 
\label{eq:hiertrip}
\centering
\begin{split}
    emb_{tax} &= f_{emb}(out_{feat}) \\
    emb_{gen} &= g_{emb}(out_{feat}) \\
    emb_{fam} &= h_{emb}(out_{feat}) 
\end{split}
\end{equation}

\begin{equation}
\centering
    \label{trip_loss}
    \begin{split}
    \mathcal{L} = \mathcal{L}_{CE_\text{tax}} +  
    \lambda_1 \mathcal{L}_{CE_\text{gen}} +   
    \lambda_2 \mathcal{L}_{CE_\text{fam}} + 
    \lambda_3 \mathcal{L}_{Trip_{\text{tax}}} +  
    \lambda_4 \mathcal{L}_{Trip_{\text{gen}}} +   
    \lambda_5 \mathcal{L}_{Trip_{\text{fam}}}
    \end{split}
\end{equation}

\paragraph{Phylogenetic Distance}
Herbarium-2022 dataset also comes with phylogenetic distances ($d_{phy}$) between two genus. We used this distance while training the network to bring closer similar genera and send apart different genera. In this experiment, we took additional genera embedding head ($\in \mathbb{R}^{128}$) from the network while using the hierarchical labels of genera, family and category (equation~\ref{doub_loss}). 

\begin{equation}
    \label{eq:dist}
    \mathcal{L}_{\text{dist}} = \left|\left| \left( \sum_{i=1}^{n}|x_i^p| \right)^{\frac{1}{p}} -d_{phy} \right|\right|^2
\end{equation}

\begin{equation}
    \label{doub_loss}
    \mathcal{L} = \mathcal{L}_{CE_\text{cat}} +  
    \lambda_1 \mathcal{L}_{CE_\text{gen}} +   
    \lambda_2 \mathcal{L}_{CE_\text{fam}} + \lambda_3 \mathcal{L}_{\text{dist}}
\end{equation}

\subsection{iNaturalist 2019}
In addition to the two datasets mentioned above, we evaluate the performance of our model on another fine-grained classification task, iNaturalist 2019. We compared the performance with other baselines involving Vision Transformers such as ConViT \cite{d2021convit}, Masked Auto Encoders (MAE) \cite{he2022masked}, Convolution-enhanced image Transformer (CeiT) \cite{yuan2021incorporating}, LeViT \cite{graham2021levit}. For training \textit{Conviformer}, we followed the default hyperparameter settings with CE loss for 300 epochs.

\paragraph{Compatibility}
This part explains how to load the base architecture pre-trained weights to our proposed model. Once we pass the input image through the convolutional block, we obtain feature vector $x$ $\in \mathbb{R}^{t_p \times 64}$. In a general scenario, an input image is of resolution 224, and $t_p$ is 196. So we remove the patch embedding projection weight and bias from the state dictionary and load the weight. To go from \textit{Conviformer} to ConViT or any other Vision transformer, we follow a similar step and remove the convolutional block weights from the patch embedding in the state dictionary and load the weights.

\paragraph{Hardware Accelerators} 
All the experiments are performed on NVIDIA RTX A5000 or Nvidia Tesla V100 GPUs using Pytorch Data Distributed Parallel (DDP) framework. In our experimental setup, we used 8 GPUs in DDP mode for training each model with a batch size of 80 per GPU for the image of resolution 448$\times$448. 

\section{Results}

\paragraph{Resolution}
This experiment shows the network's performance with increasing image resolutions. We find that as we go from 224-dimensional images to 448, there is an 8.7\% gain in accuracy. As we further increased the image's resolution to 512 and 600, performance gain increased by 5.6\% and .67\%, respectively. We did not see much improvement in extending the image resolution from 512 to 600, so we stopped. This trend shows the vast possibility of improving the accuracy in a fine-grained classification task when the networks can handle higher-resolution images. 

\begin{table}[ht]
\centering
\vspace{5mm}
\begin{tabular}{lcccc}
\cline{2-5}
\cline{2-5}
\multicolumn{1}{l}{} & \multicolumn{4}{c}{\cellcolor{LightGrey}\textbf{Image Resolution}} \\ 
\multicolumn{1}{c}{}  & \multicolumn{1}{c}{\cellcolor{LightGrey}224} & \multicolumn{1}{c}{\cellcolor{LightGrey}448} & \multicolumn{1}{c}{\cellcolor{LightGrey}512} & \multicolumn{1}{c}{\cellcolor{LightGrey}600} \\ \hline
\multicolumn{1}{l}{\cellcolor{LightGrey} \textbf{Architecture}} & \multicolumn{1}{c}{ConViT} & \multicolumn{3}{c}{\textit{Conviformer}} \\ 
\cellcolor{LightGrey} \textbf{Accuracy}  &     69.92   &  75.99 &    80.21 &   80.74       \\\hline 
\end{tabular}
\vspace{5mm}
\caption{\textbf{iNaturalist 2019}: Accuracy obtained for different input image resolutions while training the model with random initializing.}
\vspace{5mm}
\label{tab:img_res_acc}
\end{table}

\subsection{Herbarium 2021}
We trained the model for 300 epochs. During training, our model could achieve a validation accuracy of 98.3\%. We evaluated the model trained on Herbarium 2021 with \textit{conviformer} on the test images used for ranking on the Kaggle platform. Our model obtained an $\mathcal{F}1$ score (public) of \textbf{0.729} using a single model training without any ensembling methods on images of resolution 448$\times$448. This score is also the \textit{best} uni-model score reported for the challenge \cite{herbarium2021}. We also believe that other teams would have used some ensembling methods while submitting their prediction file; our model's performance still excels those. We have summarized the results in Table~\ref{tab:img_res_herb21}.

\begin{table}[ht]
\centering
\vspace{5mm}
\begin{tabular}{lcc}
\hline
\rowcolor{LightGrey} \textbf{Architecture} & \textbf{Resolution} & \textbf{$\mathcal{F}1$ score} \\ \hline
SE ResNeXt-101$^*$ \cite{hu2018squeeze}  &  448 &   .726  \\
SE ResNeXt-50$^*$ \cite{hu2018squeeze} & 448 & .708 \\
DeIT (B) \cite{touvron2021training} & 224 & .706 \\
\rowcolor{LemonChiffon} \textit{Conviformer} (Ours)  & 448 & \textbf{.729}    \\ \hline      
\end{tabular}
\vspace{5mm}
\caption{\textbf{Herbarium 2021}: $\mathcal{F}1$ score on the test set of Herbarium 2021 obtained with different architectures. We also mention the input image resolution used in training the network. ($^*$ means scores are taken from the public leaderboard of the Kaggle competition).}
\vspace{5mm}
\label{tab:img_res_herb21}
\end{table}

\subsection{Herbarium 2022}
\paragraph{Cross-entropy (CE) loss} In the first setup of the experiment, we only used the category labels and trained the network with a CE loss. The validation accuracy of our trained model achieved nearly 1.0 $\mathcal{F}1$ score. This dataset is taken as a subset of the training dataset using stratified sampling. Training the model using 90\% of the data leads to a model with an $\mathcal{F}1$ score of .814 on the test set. Later we merged the validation set to the training dataset to fine-tune the model using all available training samples for 10 more epochs. This increased our $\mathcal{F}1$ score to .824 (summarized in Table~\ref{tab:res_trip}). We ran additional training where the dropout is set at 10\% instead 0 as set by default and found a 1.5\% gain in $\mathcal{F}1$ score.

\begin{table}[ht]
\centering
\begin{tabular}{lcc}
\hline
\rowcolor{LightGrey} \multicolumn{1}{c}{\textbf{Steps}} & \textbf{$\mathcal{F}1$ score} & Epochs\\ \hline
\begin{tabular}[c]{@{}l@{}} (a) Trained on 90\% dataset with CE loss\end{tabular}         & 0.814  & 300           \\ \hline
\begin{tabular}[c]{@{}l@{}} (b) Trained on full dataset with CE loss\end{tabular}      & 0.824  & +10           \\ \hline
\begin{tabular}[c]{@{}l@{}} (c) Training on full dataset with Triplet regularization\end{tabular} & 0.827   & +10       \\ \hline
\begin{tabular}[d]{@{}l@{}} (c) Training on full dataset with ConViT~\cite{d2021convit} and CE loss \end{tabular} & 0.802  & 300           \\ \hline
\end{tabular}
\vspace{5mm}
\caption{\textbf{Herbarium 2022}: $\mathcal{F}1$ score obtained on the test dataset following different strategies. We obtain a gain of 1.5\% in $\mathcal{F}1$ score when the model is trained on the full dataset with cross-entropy (CE) loss and further regularization using triplet loss.}
\vspace{5mm}
\label{tab:res_trip}
\end{table}

At this stage, the $\mathcal{F}1$ score on the dataset without using \textit{PreSizer} is 0.813, which shows how our pre-processing technique helps get us an overall gain in the final score. Finally, we further fine-tuned this fully trained model with triplet regularization. Our triplet loss trained model gave us an $\mathcal{F}1$ score of .827. We have summarized the results in Table.~\ref{tab:res_presizer}. 

\begin{table}[ht]
\centering
\vspace{5mm}
\begin{tabular}{lc}
\hline
\rowcolor{LightGrey} \multicolumn{1}{c}{\textbf{Steps}}                                                            & \textbf{$\mathcal{F}1$ score} \\ \hline
Without \textit{PreSizer}       & 0.813             \\ \hline
With \textit{PreSizer}        & \textbf{0.824}             \\ \hline

\end{tabular}
\vspace{5mm}
\caption{\textbf{Herbarium 2022}: $\mathcal{F}1$ score obtained on the test dataset with and without using our novel augmentation technique \textit{PreSizer}. The model is trained on a full dataset using cross-entropy loss.}
\vspace{5mm}
\label{tab:res_presizer}
\end{table}

\paragraph{Hierarchical loss} 
For the next part of the experiment, we used hierarchical labels provided with the dataset. We followed a similar procedure as mentioned above to train the network. We trained the network for 300 epochs in their default training setup. When evaluated on the test dataset, it yielded a $\mathcal{F}1$ score of \textbf{.829}. We also compared the network's performance when trained from scratch, i.e., with randomly initialized weights. $\mathcal{F}1$ score when trained from scratch is 0.822. This score shows that a model using hierarchical labels, when trained from scratch, could reach the model's performance with a strong initialization, i.e., pre-trained on Herbarium data. While training, both $\lambda_1$ and $\lambda_2$ are set as 1.

\paragraph{Hierarchical + Triplet loss} 
We also used triplet regularization at a hierarchical level. At this stage, we sampled images either at \textit{family} level or at \textit{genus} level. $\mathcal{F}1$ score obtained on the test dataset at both these levels are 0.825 and 0.824, respectively. It shows that sampling at the genera level is more beneficial than at the family level because there is more similarity among genera than in the family, so triplet margin loss at this level performed better than at the family level. We have summarized these results in Table~\ref{tab:resulthier}. We set $\lambda_i=1$ $\forall i \in [1,5]$ and compute the \textit{euclidean} distance between the anchor and positive/negative class in the triplet loss.

\begin{table}[]
\centering
\begin{tabular}{lc}
\hline
\rowcolor{LightGrey} \multicolumn{1}{c}{\textbf{Steps}}                                                       & \textbf{$\mathcal{F}1$ score} \\ \hline
\begin{tabular}[c]{@{}l@{}} (a) Trained with CE loss at Family,  Genus \\ and Category level (pre-trained weights) \end{tabular}  & \textbf{0.829 }           \\ \hline
\begin{tabular}[c]{@{}l@{}} (b) Trained with CE loss at Family, Genus \\ and Category level (from scratch) \end{tabular}      & 0.822             \\ \hline
\begin{tabular}[c]{@{}l@{}} (c) Training with Triplet regularization at \\ Genus level and CE loss at Family, Genus \\ and Category level \end{tabular} & 0.825            \\ \hline
\begin{tabular}[c]{@{}l@{}} (d) Training with Triplet regularization at \\ Family level and CE loss at Family, Genus \\ and Category level \end{tabular} & 0.824          \\ \hline
\end{tabular}
\caption{\textbf{Herbarium 2022}: $\mathcal{F}1$ score obtained on the test dataset following regularization with hierarchical information. We found that the model trained using cross-entropy (CE) loss at hierarchical levels performs best, and there is no further need for triplet regularization.}
\vspace{5mm}
\label{tab:resulthier}
\end{table}

\paragraph{Phylogenetic Distance} 
In our final experiment, we used the phylogenetic distance between different genera and constrained the network to process the embedding distance between those genera similarly. This loss term is added together in addition to the hierarchical training. At this stage, we obtained the $\mathcal{F}1$ score on the test dataset as 0.824. We set $\lambda_1=\lambda2=1$ and $\lambda_3=0.1$ and $p=2$ to compute the \textit{euclidean} distance between the two vectors.

\paragraph{Phylogenetic Distance} 
In our final experiment, we used the phylogenetic distance between different genera and constrained the network to process the embedding distance between those genera similarly. This loss term is added together in addition to the hierarchical training. At this stage, we obtained the $\mathcal{F}1$ score on the test dataset as 0.824. We set $\lambda_1=\lambda2=1$ and $\lambda_3=0.1$ and $p=2$ to compute the \textit{euclidean} distance between the two vectors.

\paragraph{Final Recipe}
Finally, we want to share the final ingredients we learned after training different models for several hours on the computing clusters. First and foremost is to go higher in terms of image resolution. Our analysis shows that a 512-dimensional image could give the best performance and is a good tradeoff with the computational cost associated with training. Next, we suggest making use of hierarchical labels if available. We found these hierarchical labels to eliminate the effect of initialization. Our experimentation also found that training with the hierarchical labels performed equally to a model trained with contrastive learning. Additionally, we found a positive effect of using a dropout of 10\%, which was otherwise 0 in the default parameters. In the end, we recommend using the smart resizing technique where the images do not have a symmetric dimension. 

\paragraph{Top entry model} We used ensembling techniques like balancing predicted classes according to the probability score obtained across multiple random cropped images. We also added a dropout of 10\% while training. With these methods, we obtained the final $\mathcal{F}1$ score of 0.861 and \textbf{0.868} on the public and private test datasets, respectively. This score is amongst the top 5 entries on the Kaggle leaderboard. 

\subsection{iNaturalist 2019}

In Table \ref{tab:img_res_inat19} we have summarized the results obtained by the \textit{Conviformer} and other baselines. Our model could surpass other baselines reported on the dataset and obtain an accuracy of 82.85\%. To our knowledge, this is the best accuracy reported so far with similar parameter models. The other two better performing transformer architectures are MixMIM-L and MAE (ViT-H), which have $\sim$300M and $\sim$600M parameters and perform at 83.9\% and 88.3\%  respectively.

\begin{table}[ht]
\centering
\vspace{5mm}
\begin{tabular}{lcc}
\hline
\rowcolor{LightGrey} \textbf{Architecture} & \textbf{Resolution} & \textbf{Accuracy} \\ \hline
ConViT \cite{d2021convit}  &  224 &   77.84  \\
MAE (ViT-B) \cite{he2022masked} & 224 & 80.50 \\
MaxMIM (B) \cite{liu2022mixmim} & 224 & 82.60 \\
CeiT-S \cite{yuan2021incorporating} & 384 & 82.70 \\
LeViT \cite{graham2021levit} & 384 & 74.30 \\
\rowcolor{LemonChiffon}\textit{Conviformer} (Ours)  & 448 & \textbf{82.85}    \\ \hline      
\end{tabular}
\vspace{5mm}
\caption{\textbf{iNaturalist 2019}: Accuracy obtained with different baselines. We also mention the input image resolution used in training the network.}
\vspace{5mm}
\label{tab:img_res_inat19}
\end{table}

\paragraph{Compatibility}
We evaluated the compatibility of the weights trained on the baseline architecture (in this case, ConViT) with \textit{Conviformer}. To check this, we compared the model trained with randomly initialized weights and ImageNet pre-trained weights available on the PyTorch library. We obtained a considerable performance difference of 6.86\% in accuracy between the two (Table~\ref{tab:inat_back}). It shows that when training the \textit{Conviformer}  initialized with the pre-trained weight of the baseline model, it even outperforms the baseline. 

\begin{table}[ht]
\centering
\vspace{5mm}
\begin{tabular}{lcc}
\cline{2-3}
\cline{2-3}
\multicolumn{1}{l}{} & \multicolumn{2}{c}{\cellcolor{LightGrey}\textbf{Initialization}} \\ 
\multicolumn{1}{c}{}  & \multicolumn{1}{c}{\cellcolor{LightGrey}Random} & \multicolumn{1}{c}{\cellcolor{LightGrey}ImageNet} \\ \hline
\textbf{ConViT}  &   69.92 & 77.84 \\\hline
\rowcolor{LemonChiffon} \textbf{Conviformer}  &  75.99 &  82.85       \\\hline
\end{tabular}
\vspace{5mm}
\caption{\textbf{iNaturalist 2019}: Accuracy obtained when the \textit{Conviformer} is initialized with random weights and ImageNet pre-trained weights of ConViT. There is a gain of 9\% in the accuracy when we load the ImageNet weights to the network representing compatibility with weights trained on the baseline model.}
\vspace{5mm}
\label{tab:inat_back}
\end{table}

\section{Conclusion}

This paper put forward an elegant method helpful for fine-grained classification challenges such as Herbarium-202x and iNaturalist 2019. We first demonstrated the importance of resolution for fine-grained classification tasks. When increasing the image resolution from 224 to 512, we found an improvement of 14.72\% in accuracy on the iNaturalist 2019 dataset. Our pre-processing method, \textit{PreSizer}, resize the images without changing their aspect ratio while infusing more meaningful information using reflective padding and removing the artifacts around bordering pixels. This augmentation method helped to improve the $\mathcal{F}$1 score by 1.4\% on the Herbarium 2022 test dataset. We introduced an architecture \textit{Conviformer}, which enables a vision transformer to use higher dimensional images. Our evaluation with ConViT as base architecture exhibited an improvement of 2.7\% on test $\mathcal{F}$1 score. We also demonstrated its compatibility with the base architecture to initialize with the pre-trained weights and use the default hyper-parameter setup. This initialization is critical in training any transformer architecture. We did a robust analysis using meta-information provided with the Herbarium-2022 dataset, like hierarchical labels and phylogenetic distances. We found that hierarchical labels are essential to achieving higher accuracy in a fine-grained classification task. Overall, we proposed a generic solution that can be used for several computer vision tasks.

\section{Acknowledgement}
This work was funded by NSF (IIS-1912280) and ONR (N00014-19-1-2029) to TS and ANR (OSCI-DEEP grant ANR-19-NEUC-0004) to RV. Additional support was provided by the ANR-3IA Artificial and Natural Intelligence Toulouse Institute (ANR-19-PI3A-0004), the Center for Computation and Visualization (CCV) and High-Performance Computing (HPC) resources from CALMIP (Grant 2016-p20019, 2016-p22041). We acknowledge the Cloud TPU hardware resources that Google made available via the TensorFlow Research Cloud (TFRC) program and computing hardware supported by NIH Office of the Director grant S10OD025181.


{\small
\bibliographystyle{unsrt}
\bibliography{main}
}






\end{document}